\documentclass{article}

\usepackage[preprint]{neurips_2024}

\usepackage[utf8]{inputenc} 
\usepackage[T1]{fontenc}    
\usepackage{hyperref}       
\usepackage{url}            
\usepackage{booktabs}       
\usepackage{amsfonts}       
\usepackage{nicefrac}       
\usepackage{microtype}      
\usepackage{xcolor}         
\usepackage{graphicx} 
\usepackage{multirow}
\usepackage{authblk}

\usepackage{amsthm}
\usepackage{amsmath} 
\newtheorem{prop}{Proposition}

\newtheorem{lemma}{Lemma}
\newtheorem{corollary}{Corollary}

\title{A Markov Random Field Multi-Modal Variational AutoEncoder}

\author[1,2]{Fouad Oubari} 
\author[2]{Mohamed El Baha}
\author[2]{Raphael Meunier}
\author[2]{Rodrigue Décatoire}
\author[1,3]{\mbox{Mathilde Mougeot}}

\affil[1]{Centre Borelli, UMR 9010, ENS Paris Saclay}
\affil[2]{Michelin}
\affil[3]{ENSIIE}

\affil[ ]{\texttt{\{fouad.oubari, mmougeot\}@ens-paris-saclay.fr} \\
\texttt{\{\mbox{mohamed.el-baha, raphael.meunier, rodrigue.decatoire}\}@michelin.com}}

\begin{document}

\maketitle

\begin{abstract}
  Recent advancements in multimodal Variational AutoEncoders (VAEs) have highlighted their potential for modeling complex data from multiple modalities. However, many existing approaches use relatively straightforward aggregating schemes that may not fully capture the complex dynamics present between different modalities. This work introduces a novel multimodal VAE that incorporates a Markov Random Field (MRF) into both the prior and posterior distributions. This integration aims to capture complex intermodal interactions more effectively. Unlike previous models, our approach is specifically designed to model and leverage the intricacies of these relationships, enabling a more faithful representation of multimodal data. Our experiments demonstrate that our model performs competitively on the standard PolyMNIST dataset and shows superior performance in managing complex intermodal dependencies in a specially designed synthetic dataset, intended to test intricate relationships.
\end{abstract}

\section{Introduction}

Dealing with multimodal datasets is an essential and challenging task in modern machine learning research. These datasets integrate heterogeneous data sources, offering a comprehensive view necessary for robust and versatile models. The complexity of multimodal data extends to diverse domains such as healthcare \citep{kline2022multimodal,mohsen2023scoping}, multimedia \citep{girdhar2023imagebind,zhu2019multi}, finance \citep{xie2024pixiu, lee2020multimodal}, and industry \citep{cobb2023aircraftverse, 10.1007/978-3-031-62281-6_17}. Developing robust and expressive multimodal generative models requires accurately capturing the complex dependencies and interactions between different modalities.

Multimodal Variational AutoEncoders are central in generative modeling, enabling the learning of latent representations that can generate complex multimodal data distributions. However, traditional multimodal VAEs often operate under assumptions that oversimplify the prior and posterior distributions. This approach fails to capture the full complexity of intermodal dependencies.
To address this limitation, we introduce in this paper the MRF MVAE, a novel Multimodal Variational Autoencoder that integrates Markov Random Fields into both the prior and posterior distributions. The capability of MRFs to effectively model complex dependencies makes them particularly suitable for capturing intricate intermodal relationships \citep{koller2009probabilistic}.

Our contributions can be summarized as follows:

\paragraph{Development of the MRF MVAE} We propose a novel multimodal Variational Autoencoder that integrates Markov Random Fields into its prior and posterior distributions.
\paragraph{The GMRF MVAE}
At the heart of our framework is the GMRF MVAE, which employs Gaussian Markov Random Fields. This foundational model supports two innovative extensions that are designed to enhance its flexibility and potentially broaden its application scope.
\paragraph{Extended Variants}
 
\begin{itemize}
    \item ALMRF MVAE: This extension employs an Asymmetric Multivariate Laplace distribution to improve handling of skewed and heavy-tailed distributions, making it especially suitable for applications in finance and biology \citep{mittnik1999maximum, guo2017heavy, klebanov2003heavy}.
    \item NN-MRF MVAE:  A hybrid model that combines a GMRF posterior with a prior modeled by neural-network-learned MRF potentials, proposing a novel integration that aims to increase flexibility and adaptability of our proposed framework.
\end{itemize}
\paragraph{Methodological Framework} Across all variants, we propose a comprehensive methodological framework for training and inference. This includes the derivation of the Evidence Lower Bound (ELBO) and customized conditional and unconditional sampling methods.
    
\paragraph{Empirical Validation} We demonstrate competitive performance on the PolyMNIST benchmark and achieve superior results on a custom copula dataset designed to evaluate intricate intermodal dependencies.

\section{Related Work}

\paragraph{Multimodal VAEs}

The field of multimodal generative models has seen substantial growth recently. Among these models, Multi-Modal Variational AutoEncoders (MMVAEs) have distinguished themselves due to their capabilities in rapid and tractable sampling \citep{vahdat2020nvae}, as well as their robust generalization performance \citep{mbacke2024statistical}. The essence of multimodal generation lies in its ability to learn a joint latent representation from multiple data modalities, encapsulating a unified distribution. Traditional MMVAE frameworks typically adopt a structure with separate encoder/decoder pairs for each modality, coupled with an aggregation mechanism to encode a cohesive joint representation across all modalities. A variety of methodologies have been introduced to synthesize these distributions within the latent space.

A foundational approach by \citep{wu2018multimodal} suggests  that the joint latent posterior can be effectively approximated through the Product of Experts (PoE) assumption. This strategy facilitates cross-modal generation at inference time without necessitating an additional inference network or a multi-stage training process, marking a significant advancement over preceding methodologies \citep{suzuki2016joint, vedantam2017generative}. However, this approach implicitly relies on the assumption that the posterior distribution can be approximated by factorisable distributions. Such an assumption presupposes independence among modalities, which may not hold true. This assumption overlooks the complex intermodal relationships intrinsic to the data, potentially limiting the model's ability to fully capture the richness of multimodal interactions. 

An alternative framework proposed by \citep{shi2019variational} employs a Mixture of Experts (MoE) strategy for aggregating marginal posteriors. This method stands in contrast to the approach used in the MVAE \citep{wu2018multimodal}, which, according to the authors, is susceptible to a 'veto phenomenon': a scenario where an exceedingly low marginal posterior density significantly diminishes the joint posterior density. In contrast, the MoE paradigm mitigates the risk associated with overly confident experts by adopting a voting mechanism among the experts, thereby distributing its density across all contributing experts. However, a critique by \citep{palumbo2023mmvae} highlights a fundamental limitation of the MMVAE approach: it tends to average the contribution of each modality. Given that the model employs each modality-specific encoder to reconstruct all other modalities, the resultant encoding is biased towards information that is common across all modalities. This bias towards commonality potentially undermines the model's ability to capture and represent the diversity inherent in multimodal datasets.


The Mixture-of-Products-of-Experts (MoPoE) framework, detailed in \citep{sutter2021generalized}, refines the aggregation methodologies previously developed by the Product of Experts (PoE) and Mixture of Experts (MoE). This approach is designed as a unification and generalization of PoE and MoE, aiming to leverage the distinct advantages of each—namely, the precise posterior approximation of PoE and the adept optimization of modality-specific posteriors by MoE. The MoPoE model is designed to enhance multimodal learning by integrating these traits.

Despite its conceptual advancements, the MoPoE model introduces a computational challenge due to its training strategy. It necessitates the evaluation of all conceivable modality subsets, which equates to $2^M-1$ training configurations for $M$ modalities. This comprehensive strategy, while beneficial for robust learning across varied modality combinations, leads to an exponential increase in computational requirements relative to the number of modalities. This aspect marks a significant limitation, especially for applications involving a large number of modalities.


To mitigate the averaging problem observed in mixture-based models, several studies \citep{sutter2020multimodal, palumbo2023mmvae} have adopted modality-specific latent spaces. Specifically, \citep{palumbo2023mmvae} identifies a 'shortcut' phenomenon, characterized by information predominantly circulating within modality-specific subspaces. To address this, an enhancement of \citep{shi2019variational}'s  model incorporates modality-specific latent spaces designed exclusively for self-reconstruction. This strategy prevents the 'shortcut' by using a shared latent space to aggregate and a modality-specific space to reconstruct unobserved modalities,  ensuring that only joint information is retained in the shared space. Despite this advancement over prior approaches by resolving the shortcut dilemma, the outlined method introduces a training procedure that encompasses both reconstruction and cross-reconstruction tasks for each modality pairing, leading to a computational requirement of $M^2$ forward passes for $M$ modalities.

\paragraph{Markov Random Fields}

Undirected Graphical Models, also called Markov Random Fields \citep{wainwright2008graphical, koller2009probabilistic, murphy2012machine}, represent a stochastic process that has its origins in statistical physics \citep{kindermann1980markov}. They were introduced to probability theory as a way to extend Markov processes from a temporal framework to a spatial one. Graphical models, including MRFs, are notoriously hard to train due to  the intractability of the partition function. This has led to numerous studies \citep{carreira2005contrastive, vuffray2020efficient, bach2002learning, tan2014learning, welling2005learning} aimed at developing more efficient methods for learning graphical models, including MRFs.

\paragraph{Markov Random Fields in Machine Learning}
In Machine Learning, Markov Random Fields have predominantly been used in image processing tasks such as image deblurring \citep{perez1998markov}, completion, texture synthesis, and image inpainting \citep{komodakis2007image}, as well as segmentation \citep{krahenbuhl2011efficient, bello1994combined}. However,
recent advancements in more efficient methodologies have led to a decline in the use of MRFs, due to the relative complexity involved in their learning processes. 

To the best of our knowledge, there are limited instances where Markov Random Fields have been integrated within generative neural networks. Among these, \citep{johnson2016composing} introduced the Structured Variational Autoencoder (SVAE), which combines Conditional Random Fields (CRFs) with Variational AutoEncoders (VAEs) to address a variety of data modeling challenges. The SVAE has been applied to discrete mixture models, latent linear dynamical systems for video data, and latent switching linear dynamical systems for behavior analysis in video sequences. This approach employs mean field variational inference to approximate the Evidence Lower Bound, targeting specific data types without explicitly focusing on intermodal relationships.

Similarly, \citep{khoshaman2018gumbolt} integrates Boltzmann Machines (BMs) as priors within VAEs, focusing on discrete variables to model complex and multimodal distributions. Their methodology suggests either factorial or hierarchical structures for the posterior distribution, aiming to effectively model complex and multimodal distributions.

Although significant advances have been made, the application of MRFs within the domain of multimodal generative models, particularly in enhancing the integration and modeling of complex dependencies among multiple modalities, remains largely unexplored. Our work seeks to bridge this gap by proposing a novel integration of MRFs within a Multimodal Variational Autoencoder framework, aimed at capturing the intricate intermodal relationships more effectively. This approach not only leverages the strengths of MRFs but also addresses the limitations observed in existing multimodal generative models.

\section{Method}
\label{seq:method}
We define $\mathbf{X} = (\text{x}_1, \dots, \text{x}_n)$ as a collection of random variables, each representing a distinct modality. Our approach employs a Multimodal Variational Autoencoder with an integrated Markov Random Field in its latent space, specifically designed to effectively capture the complex intermodal dependencies.

\subsubsection{Variational Autoencoders}
Variational Autoencoders \citep{kingma2013auto} learn a latent variable model \(p_{\theta}(\mathbf{x}, \mathbf{z}) = p(\mathbf{z})\,p_{\theta}(\mathbf{x}|\mathbf{z})\) by maximizing the marginal likelihood \(\ln p_{\theta}(\mathbf{x})\). Because the true posterior \(p_{\theta}(\mathbf{z}|\mathbf{x})\) is generally intractable, VAEs introduce a variational distribution \(q_{\phi}(\mathbf{z}|\mathbf{x})\) and maximize the evidence lower bound (ELBO):
\begin{equation}
    \label{eq:elbo}
    ELBO = \mathbb{E}_{q_{\phi}(\mathbf{z}\mid \mathbf{x})}\!\bigl[\ln p_{\theta}(\mathbf{x}\mid \mathbf{z})\bigr] 
    \;-\;
    \mathrm{KL}\!\Bigl(q_{\phi}(\mathbf{z}\mid \mathbf{x}) \,\big\|\, p(\mathbf{z})\Bigr).
\end{equation}
This objective balances a reconstruction term (the expected log-likelihood under the variational distribution) and a regularization term (the KL divergence to the prior).


\subsubsection{Markov Random Fields} provide a framework for representing joint distributions through a graphical model where, nodes represent random variables and edges represent dependencies between these variables. Mathematically, an MRF is defined over an undirected graph $G = (V, E)$ where each node corresponds to a random variable in the set $(\mathbf{z}) = \{\mathbf{z}_i\}_{i=1}^n$. The joint distribution over these random variables is specified in terms of potential functions over cliques (fully connected subgraphs) of $G$. A general mathematical definition of an MRF is given by \citep{murphy2012machine, jordan1999introduction,wainwright2008graphical,  koller2009probabilistic}:
\begin{small}
\begin{equation}
\label{eq:mrf_general_expression}
    P(\mathbf{z}) = \frac{1}{\mathcal{Z}}\exp \left[ - \sum_{C \in \mathcal{C}} \Psi_C(\mathbf{z}_C) \right]
\end{equation}
\end{small}

where $\mathcal{C}$ is the set of cliques in the graph, $\Psi_C$ are the potential functions that map configurations of the random variables within the clique to a real number, $Z_C$ denotes the set of random variables in clique $C$, and $\mathcal{Z}$ is the partition function that normalizes the distribution. In the context of our work, we model both the prior $p(\mathbf{z})$ and the posterior $q_{\phi}(Z|X_1,\cdots,X_M)$ as fully connected MRFs represented by unary $\psi_i(z_i)$ and pairwise $\psi_{i,j}(z_i,z_j)$ potentials. This leads to the specific form:
\begin{small}
\begin{equation}
\label{eq:mrf_specific}
    P(\mathbf{z}) = \frac{1}{\mathcal{Z}}\exp \left[ - \left(\sum_{i<j}^M \psi_{i,j}(\mathbf{z}_i,\mathbf{z}_j)+\sum_{i}^M \psi_{i}(\mathbf{z}_i)\right) \right]
\end{equation}
\end{small}
with $\mathbf{z} = (\mathbf{z}_1,\cdots,\mathbf{z}_M)$ . This formulation captures the dependencies between modalities in our multimodal VAE framework, leveraging the MRF's ability to model complex interactions within its latent space.

\subsection{MRF MVAE}
\label{seq:mrf_mvae}

\begin{figure}[h]
\centering
\includegraphics[width=.45\textwidth]{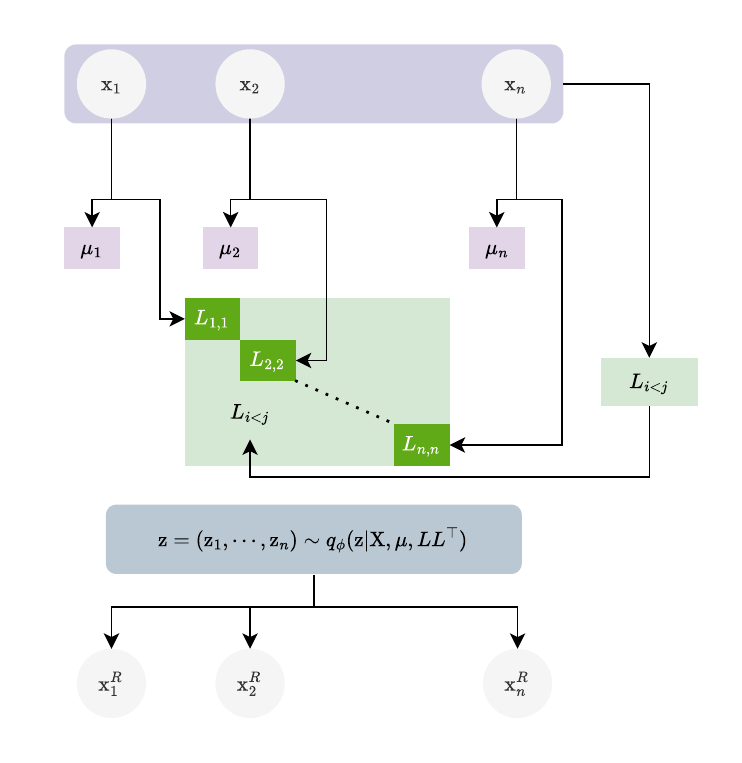} 

\caption{The MRF MVAE architecture features each encoder producing a modality-specific mean \(\mu_i\) and a diagonal block matrix \(L_{i,i}\). These matrices constitute the diagonal blocks of \(L\), the lower triangular matrix from the Cholesky decomposition of the covariance matrix \(\Sigma = LL^{\top}\). The joint posterior distribution is characterized by the concatenated mean vector \(\mu = (\mu_1,..,\mu_n)\) and the covariance matrix \(\Sigma\), with off-diagonal elements of \(L\) generated by a global encoder.}

\label{fig:mrf_mvae}
\end{figure}

While most previous works on Multimodal VAEs assume Gaussian or Laplacian distributions for the priors \( p(\mathbf{z}) \), posteriors \( q_{\phi}(\mathbf{z} \mid \mathbf{X}) \), and likelihoods \( p(\mathbf{X} \mid \mathbf{z}) \), we propose two corresponding variants (cf Figure \ref{fig:mrf_mvae}): the Gaussian Markov Random Field MVAE (GMRF MVAE) and the Asymmetric Laplace Markov Random Field MVAE (ALMRF MVAE). Additionally, we introduce a more general model where the prior is a fully connected MRF, and the potentials \(\psi_{i,j}\) and \(\psi_i\) in Equation \ref{eq:mrf_specific} are learned through neural networks. For each variant, we propose specific ELBO formulations and conditional and unconditional sampling schemes.

\subsubsection{Gaussian MRF MVAE}
In the Gaussian MRF MVAE, we assume that both prior and posterior distributions can be  approximated using a Gaussian Markov Random Field. This is characterized by a precision matrix \(\Lambda\), which dictates the dependency structure through pairwise and unary potentials. In that configuration the unary and pairwise potentials can be expressed \citep{murphy2012machine} $\psi_{i,j}(\mathbf{z}_i,\mathbf{z}_j) = \exp(-\frac{1}{2} \mathbf{z}_i^\top \Lambda_{i,j} \mathbf{z}_j)$ and  $\psi_{i}(\mathbf{z}_i) = \exp(-\frac{1}{2} \Lambda_{i,i} \mathbf{z}_i^2 + \eta_i \mathbf{z}_i)$. The joint distribution in terms of natural parameters, can be expressed as follows:

\begin{equation}
p(\mathbf{z}) \propto \exp(\mathbf{\eta}^\top \mathbf{z} - \frac{1}{2} \mathbf{z}^\top \Lambda \mathbf{z})
\end{equation}

with \(\mathbf{\eta}\) representing the natural parameter related to the mean and \(\Lambda\) the precision matrix.

In our work, we opt for the moment parameterization of the multivariate Gaussian distribution, expressed in terms of the mean vector \(\mu \in \mathbb{R}^M\) and the covariance matrix \(\Sigma\). The corresponding joint distribution is then:

\begin{small}
\begin{equation}
p(\mathbf{z}) = \frac{1}{(2\pi)^{\frac{M}{2}}|\Sigma|^{\frac{1}{2}}}\exp\left(-\frac{1}{2}(\mathbf{z}-\mu)^\top\Sigma^{-1}
(\mathbf{z}-\mu)\right)
\end{equation}
\end{small}

This representation allows for a more intuitive understanding of the distribution and simplifies certain computational aspects, such as sampling.

\paragraph{Differentiable Sampling}
The GMRF structure lends itself to differentiable sampling, a essential property for gradient-based optimization techniques in deep learning. Through the Cholesky decomposition \citep{brezinski2005methode}, we can express the covariance matrix \(\Sigma\) as the product of a lower triangular matrix \(L\) and its transpose \(L^\top\), that is \(\Sigma = LL^\top\). This factorization allows for differentiable sampling of the latent variables \citep{gentle2009computational} by first sampling a vector of i.i.d. standard normal random variables \(\mathbf{u} \sim \mathcal{N}(0, I)\) and then transforming it via \(\mathbf{z} = \mu + L\mathbf{u}\), ensuring that \(\mathbf{z} \sim \mathcal{N}(\mu, \Sigma)\) as outlined in Equation \ref{equ:reparametrization_trick_gmrf}:

\begin{equation}
\label{equ:reparametrization_trick_gmrf}
L = \text{cholesky}(\Sigma), \quad \mathbf{z} = \mu + L\mathbf{u}, \quad \mathbf{u} \sim \mathcal{N}(0, I)
\end{equation}


\paragraph{Conditional Generation}
The analytical tractability of the GMRF allows for conditional generation, which facilitates the generation of one modality given the others without additional cross-generation training. Drawing from the properties of multivariate Gaussian distributions, we have the following proposition:

\begin{prop}
\label{prop:gaussian_cond_sampling}
Given a random vector \(\mathbf{z} = (z_1, \dots, z_n) \sim \mathcal{N}(\boldsymbol{\mu}, \boldsymbol{\Sigma})\), where \(\boldsymbol{\mu} = (\mu_1, \dots, \mu_n)\) with each \(\mu_i\) of dimension \(d\), and \(\boldsymbol{\Sigma}\) is a block matrix with blocks \(\Sigma_{ij}\) of dimension \(d \times d\) representing the covariance between \(z_i\) and \(z_j\), the conditional distribution of \(\mathbf{z}_i\) given \(\mathbf{z}_j = z_j\) (for \(i \neq j\)) is Gaussian, defined as:
\begin{equation}
p(\mathbf{z}_i|\mathbf{z}_{j}=z_j) = \mathcal{N}(\hat{\mu}_i, \hat{\Sigma}_{ii})
\end{equation}
where \(\hat{\mu}_i\) and \(\hat{\Sigma}_{ii}\) are computed as:
\begin{equation}
\left\{
\begin{aligned}
\hat{\mu}_i &= \mu_i + \Sigma_{ij}\Sigma_{jj}^{-1}(z_{j} - \mu_{j}) \\
\hat{\Sigma}_{ii} &= \Sigma_{ii} - \Sigma_{ij}\Sigma_{jj}^{-1}\Sigma_{ij}^\top
\end{aligned}
\right.
\end{equation}
\end{prop}
The detailed derivation of this conditional Gaussian property is provided in Appendix \ref{app:proofs}. This significantly simplifies the conditional generation process, enabling the model to generate data for a specific modality conditioned on the observed data from other modalities without the need for explicit conditional training within the model's learning framework.

\subsubsection{Asymmetric Multivariate Laplace MRF MVAE}
\label{sec:almrf}

To enhance the robustness and fidelity of generative models in dealing with heavy-tailed skewed distributions found in domains such as finance and biology, \citep{mittnik1999maximum, guo2017heavy, klebanov2003heavy} we propose using an Asymmetric Multivariate Laplace (\(\mathcal{AL}_d\)) \citep{kotz2012laplace, kotz2001asymmetric}  distribution for both the prior and posterior distributions.

While the Gaussian Markov Random Field serves as a well-established model for capturing dependencies this section aims to extend this concept to the asymmetric Laplace distribution. Specifically, we propose the Asymmetric Multivariate Laplace Markov Random Field (ALMRF), inspired by the structural foundations of the GMRF. 

The \(\mathcal{AL}_d\) distribution, which is an extension of the symmetric Laplace distribution \citep{kotz2012laplace} characteristic function expressed as:

\begin{equation}
    \Psi(\mathbf{z}) = \frac{1}{1 + \frac{1}{2}\mathbf{z}^\top \boldsymbol{\Sigma} \mathbf{z} - \textit{i}\mathbf{\mu}^\top \mathbf{z}}
\end{equation}
This distribution provides a more flexible framework for modeling asymmetry and tail behavior in complex datasets, thereby improving the capability of multimodal generative models to accurately represent real-world phenomena.

\paragraph{Differentiable Sampling}
The Asymmetric Laplace distribution $\mathcal{AL}_d$ allows for a straightforward sampling method \citep{kotz2001asymmetric}. Let \( \text{Y} \sim \mathcal{AL}_d(\text{m}, \boldsymbol{\Sigma}) \) and \( \mathbf{X} \sim \mathcal{N}_d(\mathbf{0}, \boldsymbol{\Sigma}) \). Furthermore, consider \( W \) to be an exponentially distributed random variable with mean 1, which is independent of \( \mathbf{X} \). The random vector \( \text{Y} \) can then be generated via the transformation:
\begin{equation}
\text{Y} = \text{m}W + W^{1/2}\mathbf{X}
\end{equation}


This approach exploits the fact that the $\mathcal{AL}_d$ distribution can be conceptualized as a scale mixture of normal distributions with an exponential mixing weight.

We can sample $\mathbf{X}$ using Equation \ref{equ:reparametrization_trick_gmrf}, this provides a reparametrization trick for the ALMRF MVAE. This expression avoids incorporating the inverse of the covariance matrix $\boldsymbol{\Sigma}$ present in the density function.

\paragraph{ELBO}

To circumvent the complexities involved in the density expression of the \( \mathcal{AL}_d \) distribution within the Kullback-Leibler divergence calculation, we instead optimize the following objective function.

The proposed objective is thus:
\begin{small}
\begin{equation}
    \label{eq:almrf_vae_elbo}
    \mathcal{L} = \mathbb{E}_{q_{\phi}(\mathbf{z}|\mathbf{X})}\left[ \ln p(\mathbf{X}|\mathbf{z}) \right] - \ln \left( \text{MMD}_k^2(q(\mathbf{z}|\mathbf{X}), p(\mathbf{z})) + 1 \right)
\end{equation}
\end{small}

Here $k$ represents a c0-universal kernel. The term $\text{MMD}$ refers to Maximum-Mean Discrepancy \citep{gretton2012kernel} replacing the standard KL divergence in the ELBO formulation, reminiscent of the MMD-VAE in \citep{zhao2019infovae}. This adjustment, while analogous to MMD-VAE, ensures that the resulting formulation satisfies the following lemma proven in Appendix \ref{app:proof_lemma}:

\begin{lemma}
\label{lem:valid_elbo}
    The objective in Equation \ref{eq:almrf_vae_elbo} is a valid lower bound of \( \ln p(\mathbf{X}) \).   
\end{lemma}

\paragraph{Conditional Sampling}
Drawing from \citep{kotz2001asymmetric}, we adapt their framework to multimodal settings, enabling conditional inference within the ALMRF MVAE.

\paragraph{Distribution Nature}
The conditional generation process under the Asymmetric Multivariate Laplace ($\mathcal{AL}_d$) distribution is closely related to the Generalized Hyperbolic (GH) distribution through the following corollary:

\begin{corollary}[Generalization to \(n\)-vector Partitions]
Adapting the framework by \citep{kotz2001asymmetric}, for a random vector $\mathbf{z} = (\mathbf{z}_1, \ldots, \mathbf{z}_n) \sim \mathcal{AL}_{\sum_{i=1}^n d_i}(m, \Sigma)$, with $n \geq 2$ and each $\mathbf{z}_i$ having dimension $d_i \geq 1$, we have:
    \begin{equation}
        p(\mathbf{z}_i\mid z_j) = H_k(\lambda, \alpha, \beta, \delta, \mu, \Delta).
    \end{equation}
\end{corollary}

For the proof and detailed formulation and parameter definitions of this corollary, cf Appendix \ref{sec:proof_corollary_1}.

\paragraph{Sampling Method}

We can sample $Y$ from the GH distribution $ H_k(\lambda, \alpha, \beta, \delta, \mu, \Delta)$, using the following expression \citep{kotz2001asymmetric}:
\begin{equation}
    \text{Y} = \mu + \text{m} W+ W^{\frac{1}{2}}X
\end{equation}
where \( X \sim \mathcal{N}_d(0, \Delta) \)
and  \( W \) is a scalar random variable, independent of \(X\), following a Generalized Inverse Gaussian distribution, \( GIG(\lambda, \chi, \psi) \). The parameters' definition can be found in Appendix \ref{seq:sample_gh}.

\subsubsection{NN-MRF MVAE}
In this section, we introduce a variant within the Markov Random Field Variational Autoencoder (MRF VAE) framework, wherein the posterior is assumed to be a Gaussian Markov Random Field. To enhance the model's flexibility, we model the prior as a general MRF, employing neural networks for both unary and pairwise potentials to enrich the prior distribution \(p(\mathbf{z})\).
\paragraph{ELBO}
Our approach focuses on optimizing the ELBO. Using importance sampling, the objective function can be expressed as follows (derivation provided in Appendix \ref{app:proofs}) :
\begin{small}
\begin{equation}
    \begin{split}
        \mathcal{L} &=  \mathbb{E}_{q_{\phi}(\mathbf{z}|\mathbf{X})}\left[ \ln p(X|Z) \right] - \mathbb{E}_{q_{\phi}(\mathbf{z}|\mathbf{X})}(\ln q_{\phi}(\mathbf{z}|\mathbf{X}))\\
        &-  \mathbb{E}_{q_{\phi}(\mathbf{z}|\mathbf{X})}\left[\sum_{i<j}\psi_{i,j}^{p}(\mathbf{z}_i,\mathbf{z}_j) + \sum_{i}\psi_{i}^{p}(\mathbf{z}_i) \right] \\
        & - \ln \left[ \mathbb{E}_{q_{\phi}(\mathbf{z}|\mathbf{X})}\left(\frac{\exp \left(-\sum_{i<j}\psi_{i,j}^{p}(\mathbf{z}_i,\mathbf{z}_j) - \sum_{i}\psi_{i}^{p}(\mathbf{z}_i) \right)}{q_{\phi}(\mathbf{z}|\mathbf{X})}\right) \right]
    \end{split}
    \label{eq:elbo_nn_mrf}
\end{equation}
\end{small}
\paragraph{Conditional and Unconditional Inference}
To sample from the MRF prior, we employ the Metropolis–Hastings algorithm \citep{chib1995understanding}, which iteratively updates each variable in the latent space without requiring computation of the partition function. For conditional generation,  we fix the latent variables corresponding to known modality ($\mathbf{z}_i=z_i$) and use the Metropolis–Hastings alorithm method to estimate the remaining latent variables ($\mathbf{z}_{-i}$).

\section{Experiments}

In our empirical evaluation, we benchmark our model against four leading multimodal Variational AutoEncoders: the MVAE \citep{wu2018multimodal}, the MMVAE \citep{shi2019variational}, the MoPoE-VAE \citep{sutter2021generalized}, and the MMVAE+ \citep{palumbo2023mmvae}. The assessment focuses on two primary aspects: the quality of multimodal generation, and the models' capacity to capture complex intermodal relationships. Quality assessment is conducted on the established benchmark, the PolyMNIST dataset \citep{sutter2021generalized}.

Additionally, to assess the models' ability to capture complex intermodal dependencies, we employ a copula-based dataset. It comprises multiple uniformly-distributed modalities $(X_1,\ldots,X_M \sim \text{Uniform}(0,1))$ with interactions defined by a Gaussian copula to emulate complex relationships between modalities, such as the dependency structure found in financial markets \citep{mackenzie2014formula} or weather related events \citep{tedesco2023gaussian}. While the individual modalities appear deceptively simple, their interactions encapsulate a complexity encountered in practical applications. We average all numerical results over 3 independently trained models. 

\subsection{Assessing Generative Quality}
\label{seq:polymnist_experiment}

In this section, we evaluate the generative performance of the models using the PolyMNIST dataset.

\paragraph{The PolyMNIST Dataset :}
PolyMNIST \citep{sutter2021generalized} extends MNIST into five modalities by overlaying the same digit on different backgrounds and altering handwriting styles, creating tuples of five visually distinct yet label-consistent images. This design tests models on their ability to extract common digit identities amidst complex modality-specific settings.

\paragraph{Metrics:}
We evaluate model performance using the Fréchet Inception Distance (FID) \citep{heusel2017gans} and coherence metrics, following the methodology of Palumbo et al. \citep{palumbo2023mmvae}.  While FID is widely used to measure image similarity, our analysis (cf Section \ref{seq:qualitative_polymnist}) suggests it may not always align with human judgment, as also noted by \citep{jayasumana2023rethinking}.The Structural Similarity Index (SSIM) complements FID by assessing the perceptual quality of visual similarity between generated and real images, providing a more comprehensive evaluation.

\paragraph{Qualitative Comparison:}
\label{seq:qualitative_polymnist}

\begin{figure}[!t]
\centering
\includegraphics[width=1.\textwidth]{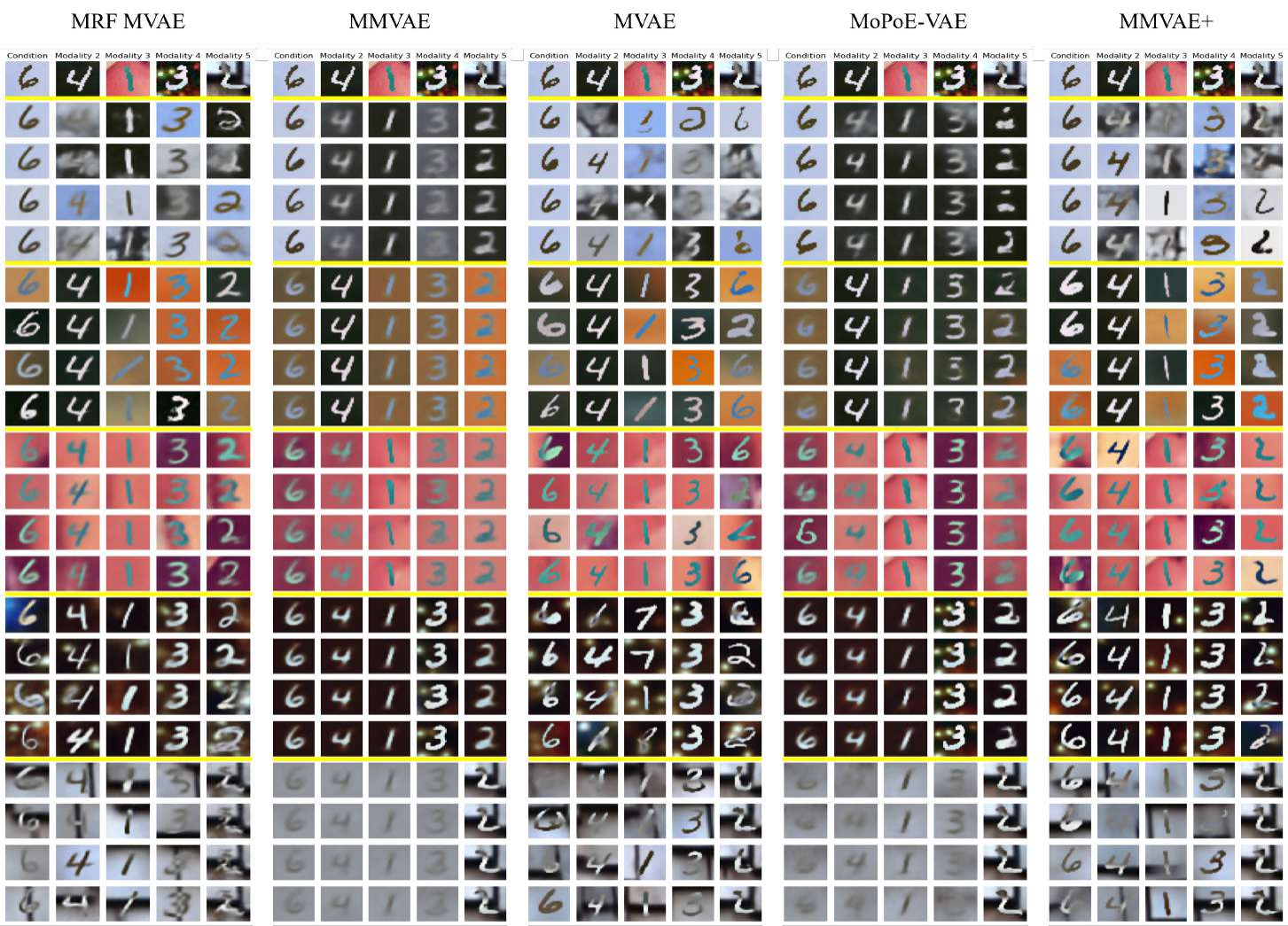} 
\caption{Illustrative comparisons of conditional sample generation using the PolyMNIST dataset. Displayed at the top row are the initial samples from one modality, followed by four samples generated conditionally for each remaining modality.}
\label{fig:conditional_generations_polymnist}
\end{figure}

We observe that The MRF MVAE model produces consistently complete digits, though slightly more blurry compared to the MMVAE+ and MVAE (cf Figure \ref{fig:conditional_generations_polymnist}) .The latter models generally deliver sharper images, but sometimes generate outputs that are incomplete or not easily recognizable as digits.

Both the MMVAE and MoPoE-VAE models struggle with diversity failing to capture varied handwriting styles and background details.

Despite some trade-offs in sharpness, our MRF MVAE consistently preserves the coherence and diversity of the digits, ensuring high generative quality and semantic coherence.

\paragraph{Quantitative Comparison:}

\begin{table}[!t]
\centering

\begin{tabular}{lcccccc}
\hline
\textbf{Model} & \multicolumn{2}{c}{\textbf{Unconditional}} & \multicolumn{3}{c}{\textbf{Conditional}} \\ \cline{2-6} 
 & \textbf{FID} & \textbf{Coherence} & \textbf{FID} & \textbf{Coherence} & \textbf{SSIM} \\ \hline
MVAE & 95.14  & 0.139  & 94.71  & 0.448  & 0.993 \\
MMVAE  & 170.87  & 0.175\normalsize & 198.80 & 0.517 & 0.995 \\
MoPoE-VAE & 106.12 & 0.018 & 162.74 & 0.475 \normalsize & 0.995 \\
MMVAE+  & \textbf{87.23} & 0.210 & \textbf{82.05} & 0.856  & 0.994 \\
MRF MVAE & 118.21 & \textbf{0.321} & 180.76 & \textbf{0.869} & 0.995 \\
\hline
\end{tabular}
\caption{Experimental results on the PolyMNIST dataset across various models and metrics.}
\label{tab:results}
\end{table}

The quantitative evaluation, as presented in Table \ref{tab:results}, reveals that the MRF MVAE model performs competitively across all considered metrics on the PolyMNIST dataset. Although the MVAE model achieves the lowest FID scores, the MRF MVAE exhibits higher values of cross-coherence and SSIM, suggesting enhanced preservation of structural integrity and global coherence in the generated samples. These results underscore the MRF MVAE's ability to produce high-quality, structurally coherent outputs, indicating its robustness in multimodal generative modeling.

\subsection{Intermodal Coherence Evaluation}
\label{sec:intermodal_coherence}

In this subsection, we evaluate the capability of each model to handle and represent complex intermodal interactions.

\paragraph{Dataset:}
Our synthetic dataset consists of four two-dimensional modalities, \(X_1, X_2, X_3, X_4\), each defined as \(X_i = (X_i^1, X_i^2)\) where each component \(X_i^j\) is uniformly distributed, \(X_i^j \sim \mathcal{U}([0,1])\), for \(i \in \{1, 2, 3, 4\}\) and \(j \in \{1, 2\}\). The coordinates of each modality are generated using two Gaussian copulas, \(C_j(X_1^j, \ldots, X_4^j)\), with uniform means \(\mu_j = [3, \ldots, 3]\) and standard deviations \(\sigma_j = [1, \ldots, 1]\). The correlation matrices \(R^j\) have off-diagonal elements set as \(R_{k,l}^j = ((-1)^j)^{k+l} \cdot 0.9\) (cf. Figure \ref{fig:joint_colpula}).

\paragraph{Metric:} We assess model performance using the Wasserstein distance \citep{villani2009wasserstein}, which measures the optimal transport cost between the empirical probability density functions (PDFs) of the generated and true samples for each modality's coordinates. This metric captures differences in both the supports and shapes of distributions. The average of these distances across all comparisons serves as an aggregate performance measure.

\subsubsection{Qualitative Comparison}

\begin{figure}[!ht]
\centering
\includegraphics[width=1.\textwidth]{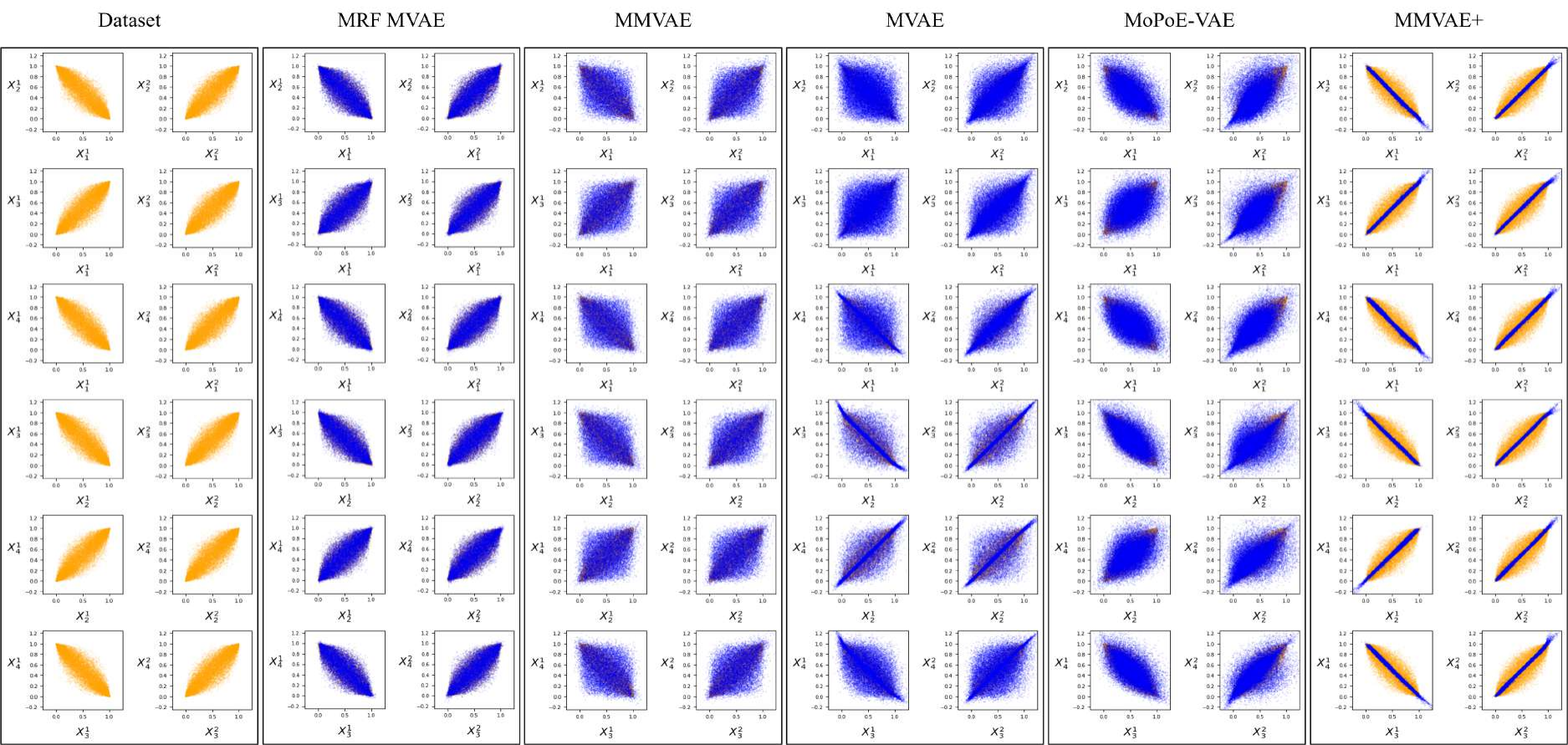} 
\caption{Qualitative results for the unconditional generations on the copula dataset. Each subplot visualizes joint distributions for each pair of coordinates \((X_i^1, X_j^1)\) and \((X_i^2, X_j^2)\) across the four two-dimensional modalities \((X_1, X_2, X_3, X_4)\). The true distributions are depicted in orange and the generated ones in blue.}
\label{fig:joint_colpula}
\end{figure}

In the analysis of joint distributions (Figure \ref{fig:joint_colpula}), the MRF VAE demonstrates superior alignment with the true distribution, indicating high-quality generations. In contrast, the MMVAE exhibits noisier and less precise outputs. The MVAE model captures certain intermodal relationships more effectively, although this varies across modalities; notably, the third modality appears less precise.
This variability is explored in further detail in appendix \ref{appendix:qualitative_results}.
As highlighted by \citep{shi2019variational}, this might be due to the "veto" effect, where experts with higher precision disproportionately impact the combined prediction. The MoPoE-VAE model integrates the distinct characteristics of both the MVAE and MMVAE, balancing sharpness and coherence effectively \citep{daunhawer2021limitations, sutter2021generalized}. Surprisingly, MMVAE+ underperforms compared to MMVAE, likely due to its increased complexity from additional modality-specific sampling, which hinders accurate representation of the true distribution's complexity.

\subsubsection{Quantitative Comparison}

\begin{table}[!h]  
\centering
\setlength{\tabcolsep}{1mm}  
\begin{tabular}{lcccccc}
\hline
\textbf{Model} & \multicolumn{3}{c}{\textbf{Uncond. Gen.}} & \multicolumn{3}{c}{\textbf{Cond. Gen.}} \\ \cline{2-7} 
 & \textbf{Dim 1} & \textbf{Dim 2} & \textbf{Mean} & \textbf{Dim 1} & \textbf{Dim 2} & \textbf{Mean} \\ \hline
MVAE & 2.7 & 3.2 & 2.9 & 3.0 & 3.1 & 3.1 \\
MMVAE & 5.2 & 4.5 & 4.8 & 5.4 & 4.7 & 5.0 \\
MoPoE-VAE & 1.9 & 2.6 & 2.2 & 6.0 & 5.6 & 5.9 \\
MMVAE+ & 8.1 & 4.9 & 6.5 & 5.2 & 4.9 & 5.1 \\
MRF MVAE & \textbf{0.7} & \textbf{0.95} & \textbf{0.86} & \textbf{2.6} & \textbf{2.7} & \textbf{2.6} \\
\hline
\end{tabular}
\caption{Comparative results on the synthetic Copula dataset for unconditional (Uncond.) and conditional (Cond.) generation using scaled Wasserstein distances (multiplied by 1000).}
\label{tab:combined_metrics_copula}
\end{table}

Table \ref{tab:combined_metrics_copula} confirms the observations from the previous section. The Wasserstein distances between the true and generated distribution PDFs indicate that the MRF MVAE generates distributions closer to the true distributions.

\section{Conclusion}
This paper integrates Variational AutoEncoders with the structural robustness of Markov Random Fields, proposing an innovative architecture that generates multiple modalities and captures complex intermodal relationships. Our MRF MVAE demonstrates competitive performance on the PolyMNIST benchmark and surpasses other state-of-the-art multimodal VAEs on the copula dataset.

Moving forward, we are committed to advancing the ELBO formulations towards a more generalized framework for all MRF MVAE variants. Additionally, we will implement sparsity enhancements in MRF potentials to boost scalability.
Furthermore, while the robust structure of MRFs inherently offers potentials that could improve explainability, we have not yet explored this possibility. Future research will focus on how these potentials can be leveraged to create more interpretable and explainable models.

\newpage

\bibliographystyle{plainnat}  
\bibliography{references}

\newpage
\appendix

\section{Technical Proofs and Derivations}
\label{app:proofs}

\subsection{Proof of proposition \ref{prop:gaussian_cond_sampling}}
\label{app:proof_gaussian_cond_sampling}

The conditional distribution of a normally distributed random variable given another is also normally distributed. This is known for the bivariate case in the Matrix Cookbook \citep{petersen2008matrix}. We extend this result for $n \geq 2$, considering a random vector $\mathbf{z} \sim \mathcal{N}(\boldsymbol{\mu}, \boldsymbol{\Sigma})$ with the following probability density function:
\begin{equation}
\mathbf{z} = 
\begin{pmatrix}
\mathbf{z}_1\\
\vdots\\
\mathbf{z}_n
\end{pmatrix}
\sim \mathcal{N}\left(
\begin{pmatrix}
\mu_1\\
\vdots\\
\mu_n
\end{pmatrix},
\begin{bmatrix} 
\Sigma_{11} & \cdots & \Sigma_{1n} \\
\vdots & \ddots & \vdots\\
\Sigma_{n1} & \cdots & \Sigma_{nn} 
\end{bmatrix} 
\right).
\end{equation}
For any pair of indices $i \neq j$ from the set $\{1, \ldots, n\}$, the conditional distribution of $\mathbf{z}_i$ given $z_j$ is
\begin{equation}
p(\mathbf{z}_i|\mathbf{z}_j= z_j) = \mathcal{N}(\hat{\mu}_i, \hat{\Sigma}_{ii}),
\end{equation}
where
\begin{equation}
\left\{
\begin{aligned}
\hat{\mu}_i &= \mu_i + \Sigma_{ij}\Sigma_{jj}^{-1}(z_j - \mu_j), \\
\hat{\Sigma}_{ii} &= \Sigma_{ii} - \Sigma_{ij}\Sigma_{jj}^{-1}\Sigma_{ji}.
\end{aligned}
\right.
\end{equation}

The following proof is inspired by a demonstration found on StackExchange\footnote{\href{https://stats.stackexchange.com/questions/30588/deriving-the-conditional-distributions-of-a-multivariate-normal-distribution}{https://stats.stackexchange.com/questions/30588/deriving-the-conditional-distributions-of-a-multivariate-normal-distribution}}.

\begin{proof}
Consider a random vector $\mathbf{z}$ and distinct indices $i$ and $j$. Define the transformation $\mathbf{y} = A \mathbf{z}_i + B \mathbf{z}_j$ such that $\mathbf{y}$ and $\mathbf{z}_j$ are independent. To achieve $\mathrm{cov}(\mathbf{y}, \mathbf{z}_i) = 0$, it follows that
\[
A \Sigma_{ij} + B \Sigma_{jj} = 0.
\]
Selecting $A = I$, leads to
\[
B = -\Sigma_{ij} \Sigma_{jj}^{-1}.
\]
Substituting back, we obtain
\[
\mathbf{y} = \mathbf{z}_i - \Sigma_{ij} \Sigma_{jj}^{-1} \mathbf{z}_j.
\]

The independence implies $\mathbf{E}[\mathbf{y}|\mathbf{z}_j] = \mathbf{E}[\mathbf{y}] = \mu_i$. Consequently, the conditional expectation of $\mathbf{z}_i$ given $z_j$ is
\begin{align*}
\mathbf{E}[\mathbf{z}_i|z_j] &= \mathbf{E}[\mathbf{y} + \Sigma_{ij} \Sigma_{jj}^{-1} \mathbf{z}_j|z_j] \\
&= \mathbf{E}[\mathbf{y}|z_j] + \Sigma_{ij} \Sigma_{jj}^{-1} z_j \\
&= \mu_i + A (\mu_j - z_j).
\end{align*}

For the variance, we derive:
\begin{align*}
\mathrm{var}(\mathbf{z}_i|z_j) &= \mathrm{var}(\mathbf{y} - B\mathbf{z}_j|z_j) \\
&= \mathrm{var}(\mathbf{y}|z_j) + \mathrm{var}(B\mathbf{z}_j|z_j) \\
&\quad - B\mathrm{cov}(\mathbf{y}, -\mathbf{z}_j) - \mathrm{cov}(\mathbf{y}, -\mathbf{z}_j)B' \\
&= \mathrm{var}(\mathbf{y}|z_j) \\
&= \mathrm{var}(\mathbf{y}).
\end{align*}

Thus:
 
\begin{align*}
\mathrm{var}(\mathbf{z}_i|z_j) &= \mathrm{var}(\mathbf{z}_i + B\mathbf{z}_j) \nonumber\\
&= \mathrm{var}(\mathbf{z}_i) + B\mathrm{var}(\mathbf{z}_j)B' + B\mathrm{cov}(\mathbf{z}_j, \mathbf{z}_i) \nonumber\\
&\quad - \mathrm{cov}(\mathbf{z}_i, \mathbf{z}_j)B' \nonumber\\
&= \Sigma_{ii} + B\Sigma_{jj}B' - B\Sigma_{ji} - \Sigma_{ij}B' \nonumber\\
&= \Sigma_{ii} - \Sigma_{ij}\Sigma_{jj}^{-1}\Sigma_{ji}
\end{align*}

This final expression for $\mathrm{var}(\mathbf{z}_i|\mathbf{z}_j = z_j)$ is the variance of the conditional distribution $p(\mathbf{z}_i|\mathbf{z}_j = z_j) = \mathcal{N}(\hat{\mu}_i, \hat{\Sigma}_{ii})$, where $\hat{\Sigma}_{ii} = \Sigma_{ii} - \Sigma_{ij} \Sigma_{jj}^{-1} \Sigma_{ji}$.
\end{proof}

\subsection{Proof of Lemma \ref{lem:valid_elbo}}
\label{app:proof_lemma}
\begin{proof}
Consider the spaces \( C(\Omega) \) and \( C_0(\Omega) \), which represent the continuous and the continuous bounded functions over a compact subset \( \Omega \subseteq \mathbb{R}^d \), respectively. Since $\Omega$ is compact,  \( C(\Omega) \) and \( C_0(\Omega) \) are the same. Let \( P \) and \( Q \) denote two multivariate $\mathcal{AL}_d$ distributions defined over this domain. Let \( C_2(\Omega, \nu) \), denote the space defined as:
\begin{equation}
    C_2(\Omega,\nu)=\{ f\in C(\Omega):\lVert f \rVert_{\nu} \leq 1 \}
\end{equation}

where $\nu$ is a regular measure and the norm operator is defined as:

\begin{equation}
    \langle f, g \rangle = \int_{\Omega} fg \, d\nu, \quad \forall f, g \in C(\Omega)
\end{equation}

To apply the inequality from Theorem 2 in \citep{wang2022practical}:
\begin{equation}
\label{equ:inequality_kl_mmd}
    KL(P \| Q) \leq \ln \left( \text{MMD}^2[ C_2(\Omega,\nu), P, Q] + 1 \right)
\end{equation}
where MMD stands for  Maximum-Mean Discrepancy. \\ We need to ensure that $P$ is absolutely continuous w.r.t $Q$. Specifically, this assumption requires that for any measurable set \( A \) within the Borel \(\sigma\)-algebra \( B(\Omega) \), if \( Q(A) = 0 \) then \( P(A) = 0 \) must also hold. Given that the density function of the $\mathcal{AL}_d$ distribution is strictly positive throughout \( \mathbb{R}^d \setminus \{0\} \), the only sets \( A \) where \( Q(A) = 0 \) could only be singleton subsets which are of measure zero. Thus, \( \forall A \in B(\Omega), Q(A)=0 \implies P(A) = 0 \). Consequently, we can use the inequality \ref{equ:inequality_kl_mmd}.

Considering that \(  C_2(\Omega,\nu) \subseteq C_0(\Omega) \), we can write
\begin{small}
\begin{align}
    KL(P \| Q) &\leq \ln \left( \text{MMD}^2[C_2(\Omega,\nu), P, Q] + 1 \right) \nonumber \\
    & \leq \ln \left( \text{MMD}^2[C_0(\Omega), P, Q] + 1 \right)
\end{align}
\end{small}

For the ELBO, we have:
\begin{small}
\begin{align}
    ELBO &= \mathbb{E}_{q_{\phi}(\mathbf{z}|\mathbf{X})}[\ln (p(\mathbf{X}|\mathbf{Z}))] - KL(q_{\phi}(\mathbf{z}|\mathbf{X}) \| p(\mathbf{z})) \nonumber \\
    &\geq \mathbb{E}_{q_{\phi}(\mathbf{z}|\mathbf{X})}[\ln (p(\mathbf{X}|\mathbf{Z}))] - \nonumber \\
    &\ln \left( \text{MMD}^2[C_0(\Omega), q_{\phi}(\mathbf{z}|\mathbf{X}), p(\mathbf{z})] + 1 \right) 
\end{align}
\end{small}

Given that $C_0(\Omega) \subset C_0(\mathbb{R}^d)$, then :

\begin{small}
\begin{equation}
    \text{MMD}^2[C_0(\Omega), q_{\phi}(\mathbf{z}|\mathbf{X}), p(\mathbf{z})] \leq \text{MMD}^2[C_0(\mathbb{R}^d), q_{\phi}(\mathbf{z}|\mathbf{X}), p(\mathbf{z})]
\end{equation}  
\end{small}


For any kernel \( k \) that is $c_0$-universal for \( \mathbb{R}^d \), the Reproducing Kernel Hilbert Space (RKHS) induced by \( k \) is dense in \( C_0(\mathbb{R}^d) \) \citep{sriperumbudur2011universality}. This means any function in \( C_0(\mathbb{R}^d) \) can be approximated arbitrarily well by functions in the RKHS.

Given this property, for any function \(f,g \in C_0(\mathbb{R}^d) \), the MMD in the RKHS induced by \(k\) : 

\begin{small}
\begin{equation}
    \text{MMD}[C_0(\mathbb{R}^d), q_{\phi}(\mathbf{z}|\mathbf{X}), p(\mathbf{z})] \approx \text{MMD}_k[q_{\phi}(\mathbf{z}|\mathbf{X}), p(\mathbf{z})]
\end{equation}
\end{small}

Since the approximation can be made arbitrarily close due to density property, we can write : 

\begin{small}
\begin{equation}
    \text{MMD}[C_0(\mathbb{R}^d), q_{\phi}(\mathbf{z}|\mathbf{X}), p(\mathbf{z})] = \text{MMD}_k[q_{\phi}(\mathbf{z}|\mathbf{X}), p(\mathbf{z})]
\end{equation}
\end{small}

Which completes the demonstration. Note that the demonstration is valid for any \( P \) absolutely continuous with respect to \( Q \), including multivariate Gaussians.

\end{proof}

\subsection{Generalization of Multivariate Asymmetric Laplace Distribution Conditional Generation}
\label{sec:proof_corollary_1}

Building upon the seminal work of \citep{kotz2001asymmetric}, we explore the generalized k-dimensional hyperbolic distribution's role in describing the conditional distribution $p(\mathbf{z}_1 \mid \mathbf{z}_2 = z_2)$, where $(\mathbf{z}_1,\mathbf{z}_2) \sim \mathcal{AL}_{r+k}(\mu, \Sigma)$, with $\mathbf{z}_1$ and $\mathbf{z}_2$ being vectors of dimensions $k, r \in \mathbb{N}^*$. This subsection aims to extend their theorem to accommodate vectors $Y=(Y_1, \ldots, Y_n)$ for $n \geq 2$.

\begin{proof}
Let \( n \geq 2 \) be a natural number, and consider a vector \( \mathbf{z} = (z_1, \dots, z_n) \) distributed according to a multivariate asymmetric Laplace distribution, \(\mathcal{AL}_{\text{ds} = \sum_{i=1}^n d_i}(\boldsymbol{\mu}, \boldsymbol{\Sigma})\), where \(\boldsymbol{\mu} = (\mu_1, \dots, \mu_n)\) and \(\boldsymbol{\Sigma}\) is defined as:
\[
\boldsymbol{\Sigma} = \begin{bmatrix}
    \Sigma_{1,1} & \cdots & \Sigma_{1,n} \\
    
    \vdots & \ddots & \vdots \\
    
    \Sigma_{n,1} & \cdots & \Sigma_{n,n}
\end{bmatrix},
\]
with each \(d_i \geq 1\) representing the dimensionality of the corresponding component \(\mathbf{z}_i\).

For any pair of distinct indices \(i, j \in \{1, \dots, n\}\) with \(i < j\) (assuming \(i < j\) without loss of generality), we define the matrix \(\mathbf{A}\) as follows:
\[
\mathbf{A} = \begin{bmatrix}
    A_{1,1} & \cdots & A_{1,\text{ds}} \\
    
    A_{2,1} & \cdots & A_{2,\text{ds}}
\end{bmatrix},
\]
where \(\mathbf{A}\) is a block matrix of dimensions \([d_i + d_j] \times \text{ds}\), with \(A_{k,l} = \mathbf{I}\) if \(k = l\) and zero otherwise. Each block \(A_{k,l}\) is a \(d_k \times d_l\) matrix.

Applying \textit{Proposition 6.8.1} from \citep{kotz2001asymmetric}, we deduce that \(\mathbf{A}\mathbf{z} = (z_i, z_j)\) follows a multivariate asymmetric Laplace distribution \(\mathcal{AL}_{d_i + d_j}(\boldsymbol{\mu}_A, \boldsymbol{\Sigma}_A)\), where:
\[
\boldsymbol{\mu}_A = \mathbf{A}\boldsymbol{\mu} = (\mu_i, \mu_j),
\]
\[
\boldsymbol{\Sigma}_A = \mathbf{A}\boldsymbol{\Sigma}\mathbf{A}^\top = \begin{bmatrix}
    \Sigma_{i,i} & \Sigma_{i,j} \\
    
    \Sigma_{j,i} & \Sigma_{j,j}
\end{bmatrix}.
\]
 which concludes the proof.
\end{proof}

Referencing \textit{Theorem 6.7.1} from \citep{kotz2001asymmetric}, the conditional probability density function is given by:

\begin{small}
\begin{equation}
\label{eq:conditional_hk}
\begin{aligned}
p(\mathbf{z}_i \mid z_j ) &= \frac{\xi^\lambda \exp(\beta^\top (\mathbf{z}_i - \mu))}{(2\pi)^{\frac{d_j}{2}}|\Delta|^{\frac{1}{2}}\delta^\lambda } \\
&\times \frac{K_{\frac{d_j}{2}-\lambda}\left(\alpha \sqrt{\delta^2 + (\mathbf{z}_i - \mu)^\top \Delta^{-1} (\mathbf{z}_i - \mu)}\right)}{K_\lambda(\delta\xi)\left[\sqrt{\delta^2 + (\mathbf{z}_i - \mu)^\top \Delta^{-1} (\mathbf{z}_i - \mu)}/\alpha\right]^{\frac{d_j}{2}-\lambda}}
\end{aligned}
\end{equation}
\end{small}

where the parameters are defined as follows:
\begin{itemize}
    \item $\lambda = 1 - \frac{d_i}{2}$
    \item $\alpha = \sqrt{\xi^2 + \beta^\top \Delta \beta}$
    \item $\beta = \Delta^{-1}(m_i - \Sigma_{i,j}\Sigma_{j,j}^{-1}m_j)$
    \item $\delta = \sqrt{z_j^\top \Sigma_{j,j}^{-1}z_j}$
    \item $\mu = \Sigma_{i,j}\Sigma_{j,j}^{-1}z_j$
    \item $\Delta = \Sigma_{i,i} - \Sigma_{i,j}\Sigma_{j,j}^{-1}\Sigma_{j,i}$
    \item $\xi = \sqrt{2 + m_j^\top \Sigma_{j,j}^{-1}m_j}$
\end{itemize}

\subsection{NN-MRF MVAE ELBO Derivation}

This section presents the derivation of the ELBO for the NN-MRF MVAE model. We begin with the standard ELBO expression:

\begin{equation}
\begin{split}
    ELBO &= \mathbb{E}_{q_{\phi}(\mathbf{z}|\mathbf{X})}\left[\ln \frac{p(\mathbf{X}, \mathbf{z})}{q_{\phi}(\mathbf{z}|\mathbf{X})}\right] \\
    &= \mathbb{E}_{q_{\phi}(\mathbf{z}|\mathbf{X})}\left[\ln p(\mathbf{X}|\mathbf{z})\right] - \mathbb{E}_{q_{\phi}(\mathbf{z}|\mathbf{X})}[\ln q_{\phi}(\mathbf{z}|\mathbf{X})] \\
    &\quad + \mathbb{E}_{q_{\phi}(\mathbf{z}|\mathbf{X})}[\ln p(\mathbf{z})]
\end{split}
\label{equ:nn_mrf_mvae_elbo}
\end{equation}

The prior \( p(\mathbf{z}) \) for the MRF model is specified as follows, incorporating both unary and pairwise potentials:

\begin{equation}
    p(\mathbf{z}) = \frac{1}{\mathcal{Z}}\exp \left[ - \left(\sum_{i<j} \psi_{i,j}(\mathbf{z}_i, \mathbf{z}_j) + \sum_{i} \psi_{i}(\mathbf{z}_i)\right) \right]
\label{eq:mrf_prior}
\end{equation}

The logarithm of the prior is given by:

\begin{equation}
    \ln p(\mathbf{z}) = -\ln \mathcal{Z} - \left(\sum_{i<j} \psi_{i,j}(\mathbf{z}_i, \mathbf{z}_j) + \sum_{i} \psi_{i}(\mathbf{z}_i)\right)
\end{equation}

The normalization constant \( \mathcal{Z} \) is computed as:

\begin{equation}
\begin{split}
    \mathcal{Z} &= \int \exp \left(-\sum_{i<j} \psi_{i,j}(\mathbf{z}_i, \mathbf{z}_j) - \sum_{i} \psi_{i}(\mathbf{z}_i)\right) d\mathbf{z} \\
    &= \mathbb{E}_{q_{\phi}(\mathbf{z}|\mathbf{X})}\left[ \frac{\exp \left(-\sum_{i<j} \psi_{i,j}(\mathbf{z}_i, \mathbf{z}_j) - \sum_{i} \psi_{i}(\mathbf{z}_i)\right)}{q_{\phi}(\mathbf{z}|\mathbf{X})}\right]
\end{split}
\label{equ:partition_function}
\end{equation}

Incorporating the expression for \( \ln p(\mathbf{z}) \) into the ELBO expression in Equation \ref{equ:nn_mrf_mvae_elbo}, we arrive at the final form of the ELBO for the NN-MRF MVAE, concluding the derivation.

\section{Distribution Characteristics and Computational Foundations}
\subsection{Density of the Asymetric Laplace Multivariate distribution}
\label{app:density_AL}
As established by Kotz et al. \citep{kotz2001asymmetric}, the density function of a d-dimensional Asymmetric Laplace \(\mathcal{AL}_d\) distributed random vector \( \text{Y} \) with mean vector \( \text{m} \) and covariance matrix \( \boldsymbol{\Sigma} \) can be expressed as:

\begin{align}
\label{equ:AL_density}
g(\mathbf{y}) &= \frac{2\exp\left(\mathbf{y}^\top \boldsymbol{\Sigma}^{-1}\text{m}\right)}{(2\pi)^{\frac{d}{2}}|\boldsymbol{\Sigma}|^{\frac{1}{2}}} \nonumber \times \left(
\frac{\mathbf{y}^\top\boldsymbol{\Sigma}^{-1}\mathbf{y}}{2+ \text{m}^\top\boldsymbol{\Sigma}^{-1}\text{m}}
\right)^{\frac{v}{2}} \nonumber \\
&\quad \times K_v\left(
\sqrt{(2 + \text{m}^\top\boldsymbol{\Sigma}^{-1}\text{m})(\mathbf{y}^\top\boldsymbol{\Sigma}^{-1}\mathbf{y})}
\right)
\end{align}

where \( v = \frac{2 - d}{2} \) and \( K_v(u) \) is the modified Bessel function of the third kind: 

$K_v(u) = \frac{1}{2} \left(\frac{u}{2}\right)^v \int_0^\infty t^{-v-1} \exp\left(-t - \frac{u^2}{4t}\right) dt, u>0$

(cf Equation \textit{A.0.4} in \citep{kotz2012laplace}).

\subsection{Sampling from Generalized Hyperbolic Distributions}
\label{seq:sample_gh}

We can sample $Y$ from the GH distribution $ H_k(\lambda, \alpha, \beta, \delta, \mu, \Delta)$, using the following expression \citep{barndorff1977exponentially}:
\begin{equation}
    \text{Y} = \mu + \text{m} W+ W^{\frac{1}{2}}X
\end{equation}
where:
\begin{itemize}
\item  $m = \Delta \beta$
\item $\mu$ and $m$ are the location and skewness parameters, respectively
\item \( X \sim \mathcal{N}_d(0, \Delta) \)
\item  \( W \geq 0 \) is a scalar random variable, independent of \(X\), following a Generalized Inverse Gaussian distribution, \( GIG(\lambda, \chi, \psi) \)
\item $\chi = \delta^{2}$, $\psi = \xi^{2}$ and $\alpha^{2} = \xi^{2} + \beta^\top \Delta \beta$
\end{itemize}

\subsection{MMD Assumption and Computation}

Formally, the MMD between the prior and posterior distributions is defined as \citep{gretton2012kernel}:

\begin{small}
\begin{align}
\label{eq:mmd_definition}
\text{MMD}_k(p, q) &= \sup_{f \in \mathcal{H}_k: \|f\|_{\mathcal{H}_k} \leq 1} \left( \mathbb{E}_{X \sim p}[f(X)] - \mathbb{E}_{Y \sim q}[f(Y)] \right) 
\nonumber\\
&= \| \mathbb{E}_{X \sim p}[k(X, \cdot)] - \mathbb{E}_{Y \sim q}[k(Y, \cdot)] \|
\end{align}
\end{small}

where $k(\cdot, \cdot)$ is the kernel function associated with the RKHS $\mathcal{H}_k$, and $\|f\|_{\mathcal{H}_k}$ denotes the norm of $f$ in $\mathcal{H}_k$.

To compute the MMD in a practical setting, we approximate it using the Monte Carlo method by sampling from both the prior and posterior distributions. The empirical estimate of MMD is given by:

\begin{small}
\begin{align}
\label{eq:mmd_empirical}
\widehat{\text{MMD}}_k(p, q)^2 &= \mathbb{E}_{X,X' \sim p} [k(X,X')] + \mathbb{E}_{Y,Y' \sim q} [k(Y,Y')] \nonumber\\
&- 2\mathbb{E}_{X \sim p, Y \sim q} [k(X,Y)]
\end{align}
\end{small}
where $X, X' \sim p$ and $Y, Y' \sim q$ are independent samples drawn from the prior and posterior distributions, respectively.

In our setting we used the universal kernel : Gaussian Radial Basis Function (RBF) kernel.

\section{Technical details for the experiments}
\label{app:technical_experiments}

Throughout all the experiments we train each model on 3 independant initializations. In this section we provide the experimental details for both PolyMNIST and the copula experiments.

\subsection{PolyMNIST Experiment}
\label{app:polymnist_experiment}

In our experiments, we employ consistent encoder/decoder architectures across all models. We use publicly available implementations for the MVAE, MMVAE, and MoPoE-VAE from \citep{sutter2020multimodal}, and for MMVAE+ from \citep{palumbo2023mmvae}. The off-diagonal elements of the covariance matrix are computed using a fully connected network with three layers, each consisting of 128 hidden units with ReLU activations, except for the final layer which uses linear activation.

To ensure the positive definiteness of the covariance matrices $\Sigma = LL^\top$ in the MRF MVAE, the diagonal elements of $L$, obtained from $\Sigma$'s Cholesky decomposition, are processed through an exponential activation function.

For all benchmark models in the experiment we configure latent spaces with dimensions of 32 for factorized models\footnote{As defined in \citep{palumbo2023mmvae}: Models with latent spaces that factorize into separate shared and modality-specific subspaces.} and 512 for unfactorized models. Each modality employs a latent space dimension of 16. To maintain uniform latent capacity across models, we apply a masking strategy that zeroes out 84\% of the off-diagonal parameters in the covariance matrices of the latent distributions, resulting in 665 parameters per distribution.

Each model was trained for up to 1,000 epochs, with the performance evaluated based on coherence and FID metrics. The best performing variants were selected for final analysis.

Specifically for the MRF MVAE models, we explored a range of  \(\beta\)\footnote{The  $\beta$ hyperparameter adjusts the weighting of the KL divergence term in the ELBO \citep{higgins2017beta}.} values: \(\{2.5 \times 10^{-3}, 1 \times 10^{-3}, 5 \times 10^{-4}, 1 \times 10^{-4}\}\). The best performing variant, as detailed in Table \ref{tab:results}, is the GMRF trained with \(\beta = 1 \times 10^{-3}\).

\subsection{Copula Dataset Experiment}
\label{app:copula_experiment}

Table \ref{tab:enc_dec_architecture} presents the architecture details for both the encoders and decoders used in the Copula experiment. To maintain consistency in latent capacities across different models, the MRF MVAE was configured with a latent dimension of 2 (yielding a total capacity of 44), whereas all other models used a latent dimension of 3 (total capacity of 48).
All models were trained for 200 epochs, exploring a range of \(\beta\) values: \(\{2.5, 1, 0.1, 0.05, 0.001\}\). For baseline models, both Gaussian and Laplacian distributions were tested for the prior, posterior, and log-likelihood calculations. Factorized and unfactorized variants were evaluated for MVAE, MMVAE, and MoPoE-VAE.

\begin{table*}[h!]
\centering
\begin{tabular}{|c|c|c|c|}
\hline
\textbf{Component} & \textbf{Layer} & \textbf{Units} & \textbf{Activation} \\
\hline
\multirow{6}{*}{Encoder} & Fully Connected & $2 \times 256$ & ReLU \\
 & Fully Connected & $256 \times 256$ & ReLU  \\
 & Fully Connected - $\sigma_{shared}$ & $256 \times latent$ & Linear\\
 & Fully Connected - $logvar_{shared}$ & $256 \times latent$ & Linear \\
\cline{2-4}
 & (if factorized) Fully Connected - $\sigma_{specific}$ & $256 \times latent$ & Linear  \\
 & (if factorized) Fully Connected - $logvar_{specific}$ & $256 \times latent$ & Linear  \\
\hline
\multirow{3}{*}{Decoder} & Fully Connected & $input \times 256$ & ReLU \\
 & Fully Connected & $256 \times 256$ & ReLU  \\
 & Fully Connected & $256 \times 2$ & Linear \\
\hline
\end{tabular}
\caption{Architecture details of the encoders and decoders used in the copula experiment.}
\label{tab:enc_dec_architecture}
\end{table*}

\section{Supplementary Experiments and Extended Results}

\subsection{Additional Results from the Copula Experiment}
\label{appendix:qualitative_results}

\paragraph{Marginal Distributions}
\begin{figure*}[!t]
\centering
\includegraphics[width=.9\textwidth]{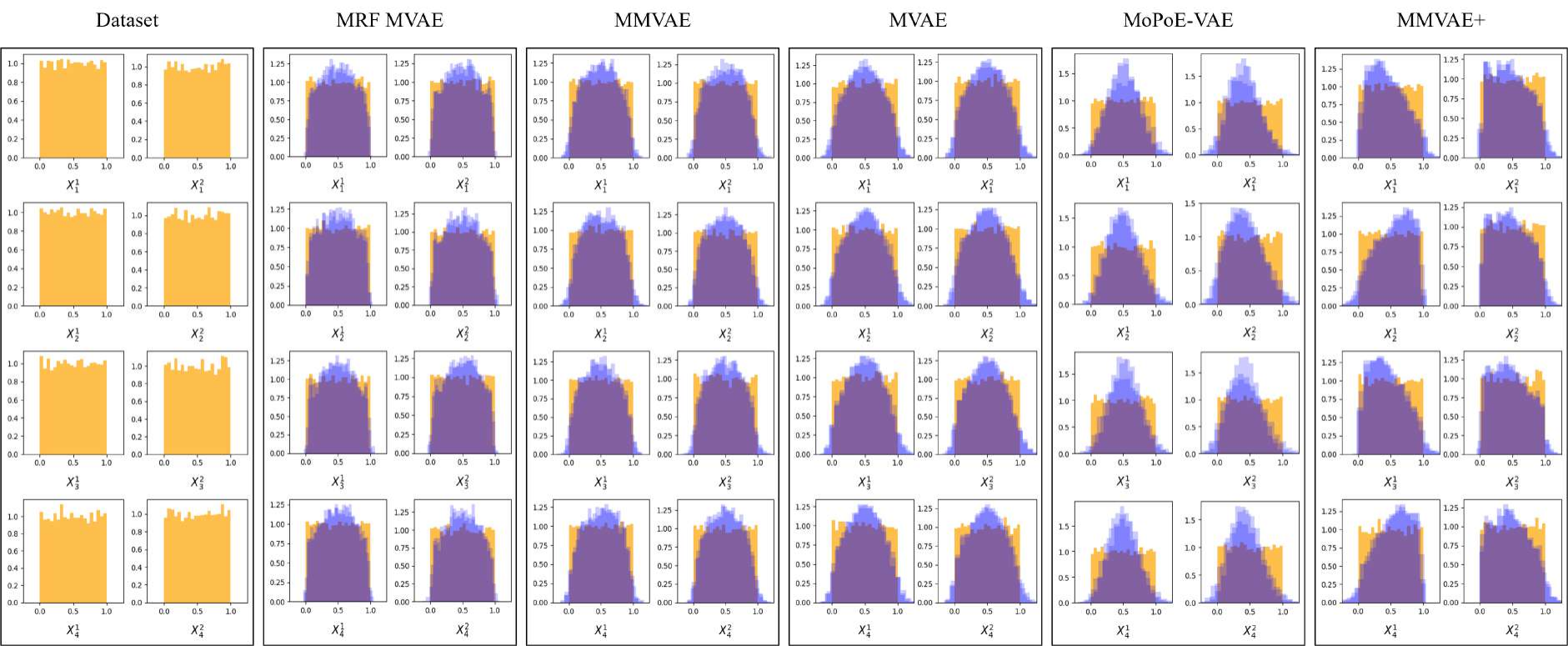} 
\caption{Qualitative analysis of unconditional generations using the copula dataset. Each subplot displays the marginal distributions for each coordinate: \((X_i^1)\) on the left and \((X_i^2)\) on the right, across four two-dimensional modalities \((X_1, X_2, X_3, X_4)\). True distributions are depicted in orange and generated distributions in blue.}
\label{fig:marginal_colpula}
\end{figure*}
As shown in Figure \ref{fig:marginal_colpula}, the marginal generations from the various baseline models and the MRF MVAE, generally conform to the expected range. Notably, the MRF MVAE closely matches the empirical marginal distributions of the dataset, consistently producing outputs within the defined range of [0,1].

\paragraph{Unconditional MVAE Generations}

\begin{figure*}[!t]
\centering
\includegraphics[width=.8\textwidth]{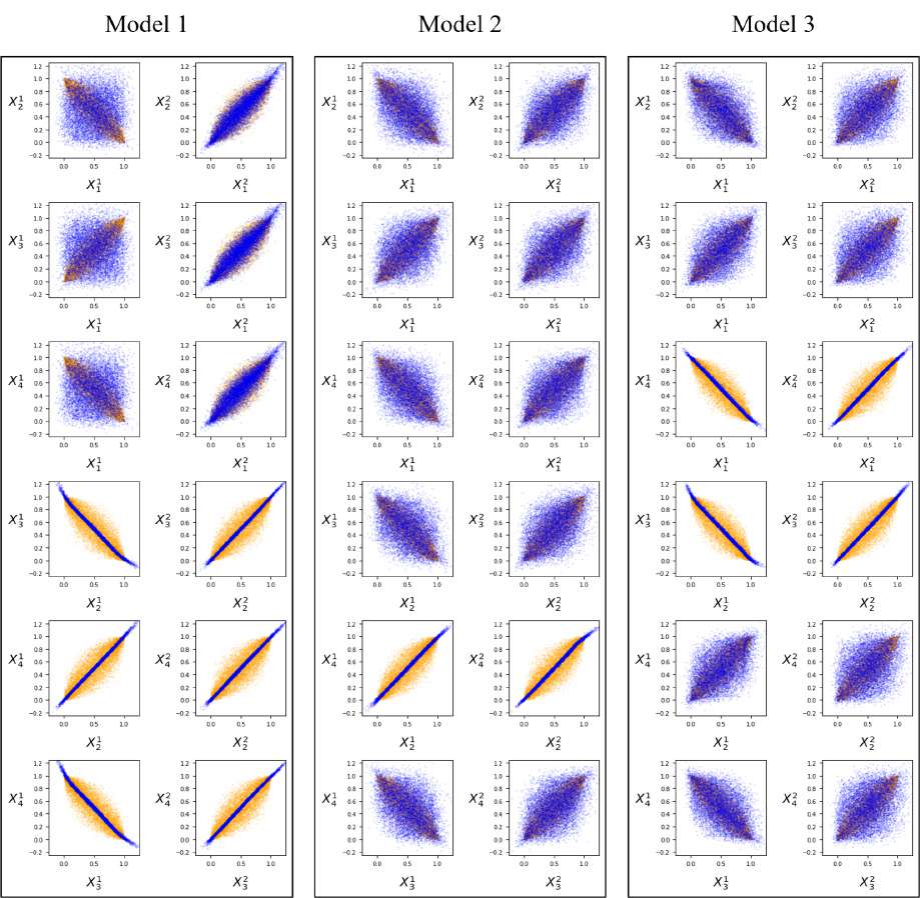} 
\caption{Qualitative results of unconditional generations from the copula dataset across three training iterations of the MVAE. Each subplot shows joint distributions for pairs of coordinates \((X_i^1, X_j^1)\) and \((X_i^2, X_j^2)\) across the four two-dimensional modalities \((X_1, X_2, X_3, X_4)\). The true distributions are shown in orange, and the MVAE-generated distributions are in blue.}

\label{fig:joint_colpula_mvae}
\end{figure*}

Figure \ref{fig:joint_colpula_mvae} displays the MVAE results after three independent trainings, revealing inconsistent alignment with the actual joint distributions between modalities. The MVAE tends to focus selectively on certain modalities, often overlooking others. This behavior reflects the "veto" effect described in \citep{shi2019variational}, where overconfident experts disproportionately influence the model's output. Such biases negatively impact the global coherence, compromising the accurate representation of intermodal relationships.

\subsection{Comparative Analysis of MRF MVAE Variants}
\label{app:mrv_variants_experiments}
This section provides a comparative analysis of the three MRF MVAE variants discussed in Section \ref{seq:mrf_mvae}. For this comparison, we employ a dataset comprising three modalities of MNIST digits. The first two modalities are similar, while the third modality is randomly selected. The training dataset consists of 60,000 examples.

\paragraph{ALMRF MVAE Implementation Details}

Sampling from the GH distribution, as proposed in Section \ref{sec:almrf}, requires sampling from the \textit{GIG} distribution. This process is known for its challenging nature and numerical instability \citep{hormann2014generating}. These instabilities are compounded by the necessity of computing matrix inversions to derive the \textit{GIG} parameters, leading to unsatisfactory conditional generations in this section. To mitigate these issues, we approximate the conditional distribution $p(\mathbf{z}_i \mid \mathbf{z}_j=z_j)$ with a Gaussian distribution that matches its mean and variance, i.e., $E(\mathbf{z}_i|\mathbf{z}_j = z_j)$ and $\mathrm{Var}(\mathbf{z}_i|\mathbf{z}_j = z_j)$.

Following \textit{Proposition 6.7.1} from \citep{kotz2001asymmetric}, the expectation and variance of $p(\mathbf{z}_i \mid \mathbf{z}_j=zj)$ can be expressed as follows:
\begin{small}
\begin{align*}
  E(\mathbf{z}_i| \mathbf{z}_j=z_j) &= \Sigma_{ij}\Sigma^{-1}_{jj}z_j \\
  &\quad + \left(m_i - \Sigma_{ij}\Sigma^{-1}_{jj}m_1\right) \frac{Q(z_1)}{C} R_{1-\frac{d}{2}}(CQ(z_j)) \\
\end{align*}  
\end{small}

\begin{small}
\begin{equation}
\begin{split}
  \mathrm{Var}(\mathbf{z}_i| \mathbf{z}_j = z_j) &= \frac{Q(z_j)}{C} \left(\Sigma_{ii} - \Sigma_{ij} \Sigma_{jj}^{-1} \Sigma_{ji}\right) R_{1-d/2}(CQ(z_j)) \\
  &\quad + (m_i - \Sigma_{ij} \Sigma_{jj}^{-1} m_j)(m_i - \Sigma_{ij} \Sigma_{jj}^{-1} m_j)^T \\
  &\quad \times \frac{Q(z_j)}{2C} G(z_j)
\end{split}
\end{equation}
\end{small}

where 
\[ C = \sqrt{2 + m_{j}^\top \Sigma^{-1}_{jj}m_j}, \]  
\[ Q(z_j) = \sqrt{z_{j}^\top \Sigma^{-1}_{jj} z_j}, \]  
\[ G(x) = R_{1-d/2}(CQ(x)) R_{2-d/2}(CQ(x)) - R_{1-d/2}^2(CQ(x)), \] 
and  
\[ R_s(x) = \frac{K_{s+1}(x)}{K_s(x)}, \] 
with $K_s$ being the modified Bessel function of the third kind \citep{kotz2001asymmetric}.

\paragraph{Architecture and Training Details}
Each of the three MRF MVAE variants employs an identical architectural configuration, featuring fully connected networks (FCNs) with four layers and 1024 hidden dimensions, alongside a latent dimensionality of 8. The covariance encoder operates through a three-layer fully connected network with 128 hidden dimensions. In the NN-MRF MVAE, potentials are derived using a three-layer, 1024-dimensional FCN that integrates concatenated latent and one-hot position encodings (corresponding to \(i,j\) for the potential \(\psi_{ij}\)). All models were trained for 100 epochs. ReLU activations are adopted for intermediate layers, with linear activations designated for outputs.

\paragraph{MRF MVAE Implementation details}

It is crucial to acknowledge that when approximating the prior distribution with a neural-network-learned MRF, the definition of the partition function, \(\mathcal{Z}\), is not guaranteed due to its integral form (cf Equation \ref{equ:partition_function}) which may not be necessarily defined. This experiment suggests that, owing to the regularization effect of the KL divergence term, the posterior distribution remains sufficiently close to the prior, ensuring the partition function's definition across the various independent runs.

\begin{figure*}[!h]
\centering
\includegraphics[width=.95\textwidth]{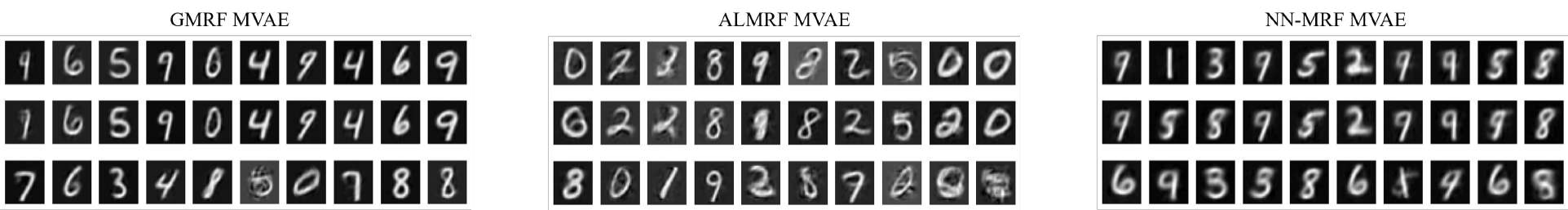}
\caption{Visual comparison of the unconditional generations across the three MRF MVAE variants.}
\label{fig:unconditional_mnist}
\end{figure*}

\begin{figure*}[!h]
\centering
\includegraphics[width=1.\textwidth]{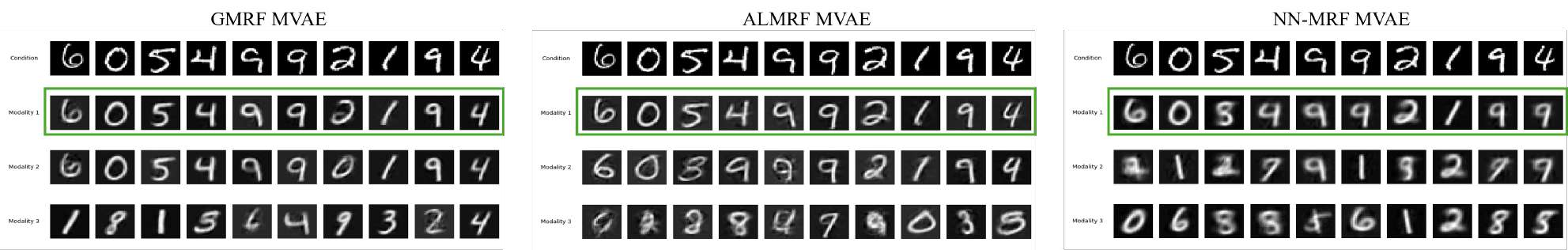} 
\caption{Visual comparison of conditional generations across the three MRF MVAE variants. The first row features test set samples, the second row shows reconstructions from the first modality, and subsequent rows present conditional generations of the remaining modalities, all conditioned on the first modality.}

\label{fig:conditional_mnist_1}
\end{figure*}
\begin{figure*}[!h]
\centering
\includegraphics[width=1.\textwidth]{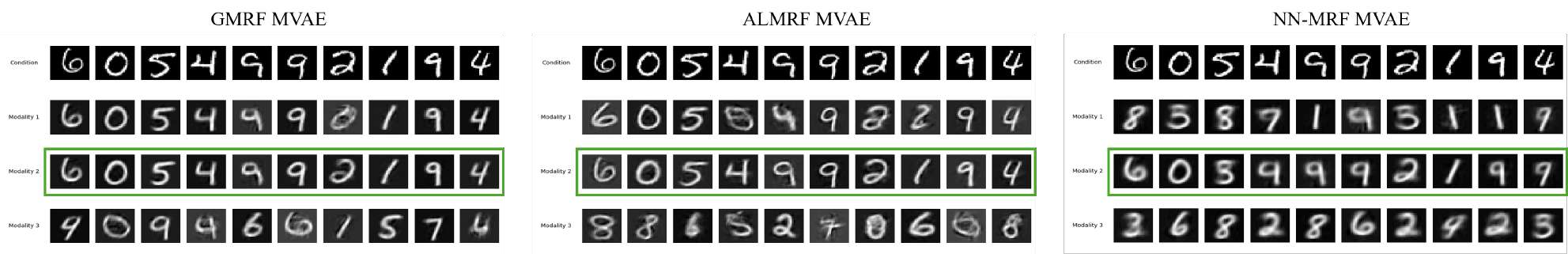} 
\caption{Visual comparison of conditional generations across the three MRF MVAE variants. The first row features test set samples, the third row shows reconstructions from the second modality, and subsequent rows present conditional generations of the remaining modalities, all conditioned on the second modality.}

\label{fig:conditional_mnist_2}
\end{figure*}

\begin{figure*}[!h]
\centering
\includegraphics[width=.65\textwidth]{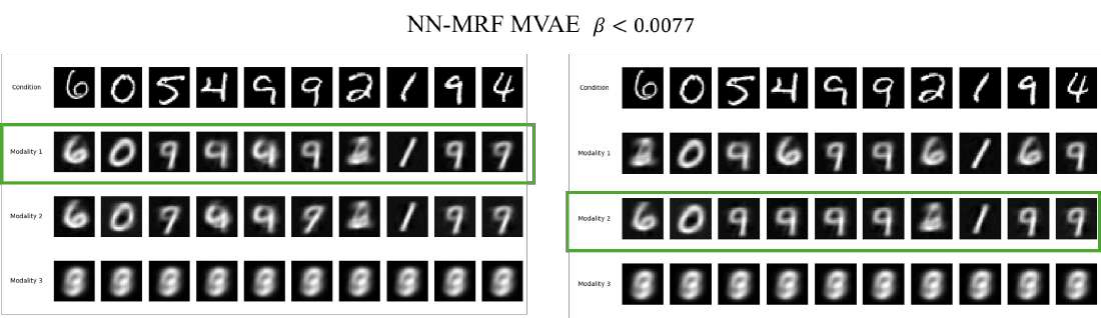} 
\caption{Conditional generations from the NN-MRF MVAE, trained with $\beta = 0.0085$. This setting enhances coherence, achieving a conditional coherence of \textbf{0.41} and an unconditional coherence of \textbf{0.57}, at the expense of the third modality's quality.}
\label{fig:cond_nn_mrf_small_beta}
\end{figure*}

\paragraph{Qualitative Evaluation}
A detailed visual analysis of the generated images, as presented in Figures \ref{fig:unconditional_mnist} and \ref{fig:conditional_mnist_1} and \ref{fig:conditional_mnist_2}, reveals notable performance differences between the model variants. Both the GMRF MVAE and ALMRF MVAE models demonstrate superior image quality and global coherence. These variants produce visually consistent and clearer images.

In contrast, the NN-MRF MVAE variant, while capable of generating visually correct images, struggles with maintaining coherence. This deficiency is particularly pronounced when the beta parameter is inadequately tuned; a larger beta ($>0.0078$) leads to coherent but noisy images in the third modality (cf Figure \ref{fig:cond_nn_mrf_small_beta}), whereas a larger beta results in clearer digits but compromised overall coherence (cf Figures \ref{fig:conditional_mnist_1} and \ref{fig:conditional_mnist_2}).

\paragraph{Quantitative evaluation}

\begin{table}[!h]
\centering

\begin{tabular}{lccccc}
\hline
\textbf{Model} & \multicolumn{2}{c}{\textbf{Uncond.}} & \multicolumn{2}{c}{\textbf{Cond.}} \\ \cline{2-5} 
 & \textbf{FID} & \textbf{Coh} & \textbf{FID} & \textbf{Coh} \\ 
 \hline
 
NN-MRF MVAE & 14252.11 & 0.57  &  16840.24 & 0.16  \\
ALMRF MVAE  & 5790.59 & 0.39 & 5489.42 & 0.76  \\
GMRF MVAE & 5320.80 & 0.82& 4418.59 &  0.94 \\

\hline
\end{tabular}
\caption{Experimental results on the MNIST dataset across conditional and unconditional FID and coherence metrics.}
\label{tab:mnidt_results}
\end{table}

The quantitative metrics presented in Table \ref{tab:mnidt_results} confirm the qualitative assessments. Both the GMRF MVAE and ALMRF MVAE models outperform the NN-MRF MVAE in terms of FID and coherence metrics under both conditional and unconditional settings.

\paragraph{Further Insights and Recommendations}
To further validate these findings and explore the full potential of each model, additional tests are recommended. Specifically, it would be beneficial to investigate the flexibility of the NN-MRF model and the ability of the ALMRF model to manage outliers and skewed data distributions. Adjustments in sampling methodologies might also be considered to enhance model performance, particularly for the NN-MRF variant.

\section{Exploring a Generalized Variant of MRF MVAE Models}

In this section, we discuss a more comprehensive configuration in which both the prior \(p(\mathbf{z})\) and posterior \(p(\mathbf{z}|\mathbf{X})\) are characterized by general Markov Random Fields. This approach opens up possibilities for robustly modeling complex intermodal relationships. However, this configuration also presents significant challenges due to the intractability of the partition functions \(\mathcal{Z}_p\) and \(\mathcal{Z}_q\), which are critical to the prior and posterior distributions. The ELBO for this model configuration is as follows:

\begin{small}
\begin{equation}
    \label{eq:mrf_elbo}
    \begin{split}
    \text{ELBO} &= \mathbb{E}_{q_{\phi}(\mathbf{z}|\mathbf{X})}[p(X|z)] - \log\left(\frac{\mathcal{Z}_p}{\mathcal{Z}_q}\right) \\
    &- \mathbb{E}_{q_{\phi}(\mathbf{z}|\mathbf{X})}\left[\sum_{i<j}\left(\psi_{i,j}^{p}(z_i,z_j) - \psi_{i,j}^{q}(z_i,z_j)\right)\right] \\
    &- \mathbb{E}_{q_{\phi}(\mathbf{z}|\mathbf{X})}\left[\sum_{i}\left(\psi_{i}^{p}(z_i) - \psi_{i}^{q}(z_i)\right)\right]
    \end{split}
\end{equation}
\end{small}

Although direct computation of \(\mathcal{Z}_p\) and \(\mathcal{Z}_q\) remains elusive, we can effectively estimate the gradient of the log partition function with respect to the model parameters (\(\theta\)) through sampling. This estimation can be expressed as follows \citep{khoshaman2018gumbolt}:
\begin{small}
\begin{equation}
    \nabla_{\theta} \ln Z_{\theta} = \nabla_{\theta} \ln \sum_{z} \exp \left(-E_{\theta}(\mathbf{z})\right) = -\mathbb{E}_{p_{\theta}(\mathbf{z})}\left[\nabla_{\theta}E_{\theta}(\mathbf{z})\right]
\end{equation}
\end{small}

where \(E_{\theta}(\mathbf{z}) = \sum_{i<j}\psi_{i,j}(z_i,z_j) + \sum_{i}\psi_{i}(z_i)\) represents the energy of configuration \(z\) under the model parameters \(\theta\). This approach enables us to navigate the partition function's intractability, facilitating the model's training through gradient-based optimization techniques.

\end{document}